%% file: celda.tex
\pdfoutput=1

\documentclass[11pt]{article}

\usepackage{ACL2023}
\usepackage{times}
\usepackage{latexsym}
\usepackage[T1]{fontenc}
\usepackage[utf8]{inputenc}
\usepackage{microtype}
\usepackage{inconsolata}
\usepackage{subcaption}
\usepackage{graphicx}
\usepackage{tabularx}
\usepackage{multirow}
\usepackage{booktabs}
\usepackage{amssymb}
\usepackage{bm}
\usepackage{mathtools}
\usepackage{relsize}
\usepackage{lipsum}
\usepackage{amsmath}
\usepackage{bibentry}

\usepackage{pgfplots}
\usepackage{tcolorbox}
\pgfplotsset{compat=1.13}
\definecolor{decentgrey}{RGB}{232,232,232}
\newtcbox{\pattern}{on line,colback=decentgrey,colframe=white,size=fbox,arc=3pt, box align=base, before upper=\strut, top=-2pt, bottom=-2pt, boxrule=0pt}

\newcommand{\txtmagenta}[1]{\textcolor{magenta}{#1}}
\usepackage[ruled,vlined]{algorithm2e}
\usepackage{cleveref}
\crefname{section}{§}{§§}
\DeclareMathOperator*{\argmax}{arg\,max}
\DeclareMathOperator*{\argmin}{arg\,min}
\newcommand{\celda}{\textsc{CeLDA}}
\newcommand{\mask}{\texttt{\txtmagenta{[MASK]}}}

\title{\celda: Leveraging Black-box Language Model as Enhanced Classifier without Labels}

\author{
    Hyunsoo Cho$^\diamondsuit$, Youna Kim$^\diamondsuit$, Sang-goo Lee$^\diamondsuit$$^\clubsuit$ \\
    $^\diamondsuit$Seoul National University, $^\clubsuit$IntelliSys\\
    \texttt{\{johyunsoo, anna9812, sglee\}@europa.snu.ac.kr}
}

\begin{document}
    \maketitle
    \input{Sections/00.Abstract}

\input{Sections/01.Introduction}

    \input{Sections/02.RelatedWork}

    \input{Sections/03.Preliminaries}

\input{Sections/04.Methods}

\input{Sections/05.Experiments}

\input{Sections/06.Analysis}
    \input{Sections/07.Conclusion}
    \input{Sections/08.Limitations}
    
    \bibliography{anthology,custom}
    \bibliographystyle{acl_natbib}

    \clearpage
    \input{Sections/99.Appendix}

\end{document}

%% file: Sections/00.Abstract.tex
\begin{abstract}

Utilizing language models (LMs) without internal access is becoming an attractive paradigm in the field of NLP as many cutting-edge LMs are released through APIs and boast a massive scale.
The de-facto method in this type of \textit{black-box} scenario is known as \textit{prompting}, which has shown progressive performance enhancements in situations where data labels are scarce or unavailable.
Despite their efficacy, they still fall short in comparison to fully supervised counterparts and are generally brittle to slight modifications.
In this paper, we propose Clustering-\textsc{e}nhanced Linear Discriminative Analysis (\celda), a novel approach that improves the text classification accuracy with a very weak-supervision signal (i.e., name of the labels).
Our framework draws a precise decision boundary without accessing weights or gradients of the LM model or data labels.
The core ideas of \celda\ are twofold:
(1) extracting a refined pseudo-labeled dataset from an unlabeled dataset, and (2) training a lightweight and robust model on the top of LM, which learns an accurate decision boundary from an extracted noisy dataset.
Throughout in-depth investigations on various datasets, we demonstrated that \celda\ reaches new state-of-the-art in weakly-supervised text classification and narrows the gap with a fully-supervised model.
Additionally, our proposed methodology can be applied universally to any LM and has the potential to scale to larger models, making it a more viable option for utilizing large LMs.

\end{abstract}

%% file: Sections/01.Introduction.tex
\section{Introduction}

    \begin{figure}[t]
    \begin{center}
        \includegraphics[width=0.99\columnwidth]{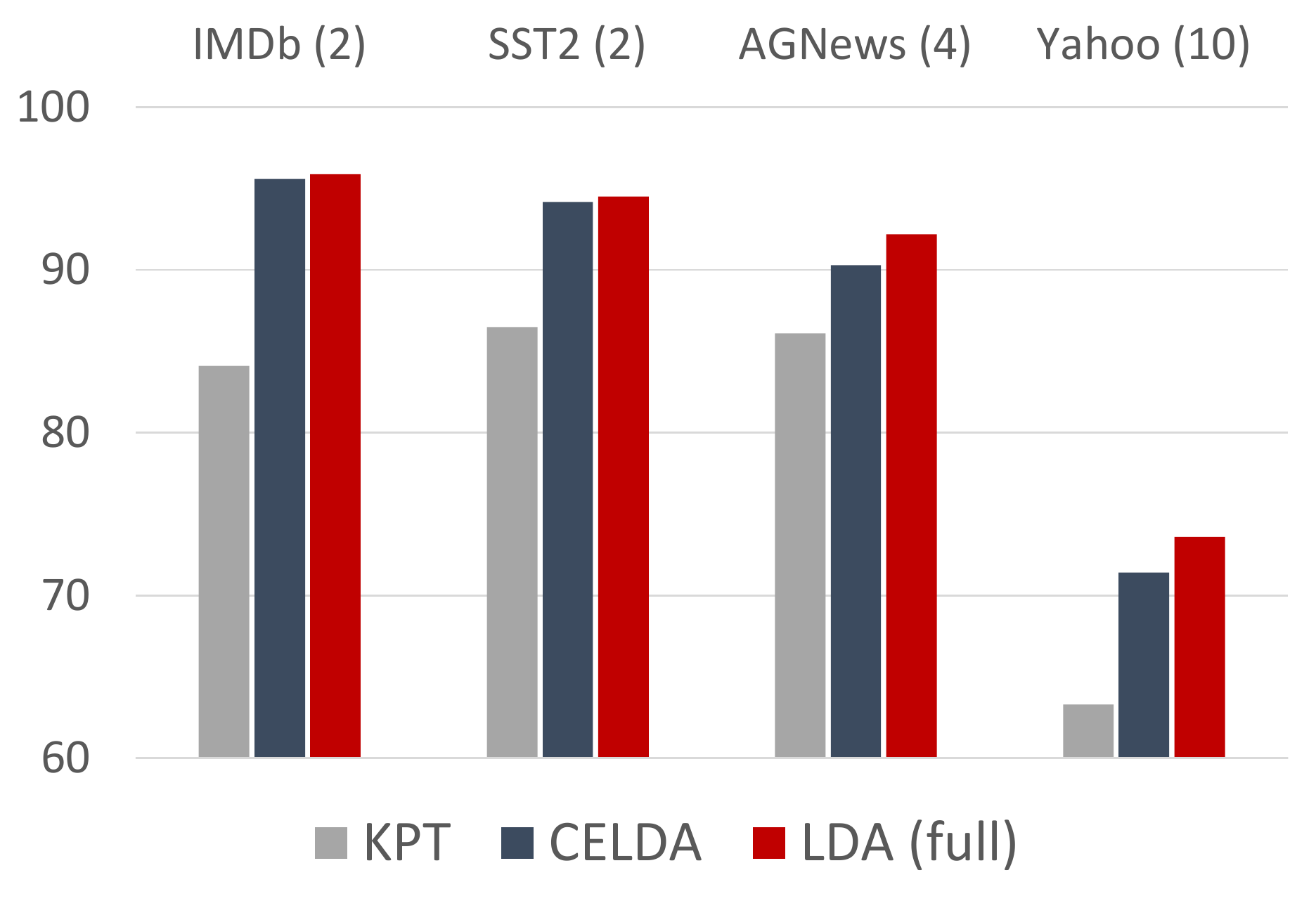}
          \caption{ 
          Text classification accuracy on 4 benchmark datasets on T5 (11B). \celda\ significantly improves the performance without model adaptation or labeled data closing the gap with fully labeled methods.
          }
          \label{fig:front_image}
    \end{center}
    \end{figure}

    Large-scale language models (LMs) have been a driving force behind a series of breakthroughs in the machine-learning community.
    Despite their preeminence in wide applications, large LMs are often costly or infeasible to fine-tune as many distributed large models, such as GPT-3 \cite{brown2020language}, are provided in a \textit{black-box} manner, which only allows limited access through commercial APIs.
    To circumvent the explicit adaptation of the models, many recent research leverage \textit{prompting}, a training-free approach that elicits desired predictions from LMs by curating input into a more conceivable form.
    By doing so, prompting has shown remarkable improvements in data-scarce scenarios (i.e., zero-shot, few-shot), reminiscing the potential of large LMs as a universal, off-the-shelf solution for diverse tasks.
    Yet, it is still premature as their performance lags far behind the fine-tuned model and displays fragility to negligible changes \cite{lu2021fantastically, perez2021true}.
    
    In this paper, we aim to bridge this gap by utilizing an unlabeled dataset without adapting the model and propose Clustering-\textsc{e}nhanced Linear Discriminative Analysis (\celda) that maximizes the potential of \textit{black-box} language models further.
    \celda\ enhances text classification performance by training a lightweight module stacked on the top of LM.
    The improvement is a result of attaining the following two key objectives:
    (1) composing a highly reliable pseudo-labeled dataset with \textit{black-box} LM.
    (2) training a compact but robust model with the previously composed dataset (pseudo-labels).
    We accomplish the first goal by sorting out some uncertain data points via clustering the features from LM. We draw inspiration from recent findings \cite{aharoni-goldberg-2020-unsupervised, cho2023prompt} that LM effectively groups semantically similar sentences into clusters.
    Furthermore, we achieve the latter objective by employing Linear Discriminative Analysis (LDA) which is efficient in terms of parameter requirements and robust against spurious inputs \cite{murphy2022probabilistic}.
    The mentioned characteristics of LDA have strong compatibility with our training dataset, considering the presence of noise and its reduced scale.
    
    To verify our method, we compare with recently proposed state-of-the-art methods on 8 different text classification datasets, spanning from binary to multi-class tasks (maximum 150 classes).
    Moreover, we also report the performance of baseline LDA when the data labels are fully available, serving as the upper bound performance.
    As illustrated in Figure 1, our method significantly outperforms the precedently proposed prompting-based zero-shot learning (ZSL) method \cite{hu2021knowledgeable} and closes the gap between fully labeled methods.
    
    In summary, our contributions are as follows:

    \noindent(1)  We propose dubbed \celda, a novel weakly-supervised learning framework for black-box language models that consistently outperforms other competing methods and close the gap between fully fine-tuned model.
    
    \noindent(2) \celda\ has the potential to scale to larger models, whereas performance often saturates in ZSL methods, and existing weakly-supervised learning methods often require additional model tuning.
    
    \noindent(3) \celda\ proves to be a highly effective active learner, capable of tackling even the most challenging tasks previously faced by ZSL or WSL methods with minimal human-in-loop labeling.

%% file: Sections/02.RelatedWork.tex
\section{Related Work}

    \noindent \textbf{Zero-shot learning} aims to identify the class label of the input sentence by relying on weak supervision information from metadata, such as the description of the task and the name of the class labels.
    ZSL methods inference test input instantly without any dataset or explicit training.
    Specifically, most ZSL works \cite{holtzman2021surface, zhao2021calibrate, min2022noisy, schick2021exploiting} utilize the likelihood of manually designed verbalizers relying on the language model's capability to predict the probability of the \mask\ word (bi-directional models) or next token (auto-regressive models).
    
    Additionally, most recent research in ZSL utilizes external knowledge bases or corpus \cite{lyu2022z, shi2022nearest, hong2022tess}. 
    \citet{lyu2022z} retrieves semantically similar samples from the additional corpus and labels them randomly.
    \citet{shi2022nearest} employs automatically expanded fuzzy verbalizers to converge the mapping between the verbalizer tokens and the class labels.
    \citet{hong2022tess} additionally uses semantically similar sentences from supplementary corpora to compensate for the poorly described labels.
    
    \noindent \textbf{Weakly-supervised learning} approaches, unlike ZSL methods, generally require an unlabeled dataset relevant to the target task and re-train the backbone model.
    Generally, most WSL studies \cite{meng2020text, wang2021x, zhang2021weakly, Fei2022BeyondPM} utilize keywords from metadata (e.g., name of each class) to annotate unlabeled datasets and iteratively re-train models with the pseudo-labeled dataset.
    
    Namely, X-class \cite{wang2021x} incrementally adds several similar words to each class until in-consistency arises and utilizes the weighted average of contextualized word representations to retrieve the most confident documents from each cluster to train a text classifier.
    SimPTC \cite{Fei2022BeyondPM} trains a Gaussian Mixture Model (GMM) on top of the LM's representations, similar to our approach in that it utilizes Gaussian distribution to fit the model.
    While effective, most WSL methods are challenging to utilize in \textit{black-box} scenarios as they often require direct model tuning or are tailored for a particular language model.

    \noindent \textbf{Black-box tuning} is a research field aiming to maximize downstream task performance without accessing weights or gradients of the model, which has diverse potential and benefits.
    Black-box tuned models can process \textit{mixed-task} input batch with a single model as it circumvents the explicit model re-training phase.
    Furthermore, it can leverage some commercial models available only through APIs \cite{brown2020language, sun2021ernie} or even when models are too large \cite{zhang2022opt, scao2022bloom} to optimize directly.
    
    In this scenario, the prevailing paradigms are: (1) manipulating the input text or (2) training a lightweight model on top of the final representations from the model.
    Specifically, the most recent work utilizes the language model's ability to learn in-context \cite{brown2020language} and tailors the input through appending templates or few-shot samples (i.e., demonstrations) to the original inputs.
    Additionally, some studies \cite{diao2022black, sun2022black} attempt to find the optimal prompt for the task without explicitly calculating the gradient.

%% file: Sections/03.Preliminaries.tex
\section{Preliminary}

\subsection{Scenario \& Problem Formulation}
    Our research aims to improve text classification without using dataset labels and accessing LM's parameters or gradients.
    Namely, LM only serves as a fixed text encoder function $\texttt{LM}(\cdot)$, which outputs an $n$-dimensional continuous latent representation $\bm{h}\in\mathcal{H}=[0,1]^{n}$ from input $\bm{x}$: $\bm{h}=\texttt{LM}(\bm{x})$.
    And let the label space be $\mathcal{Y}=\{0,1,\cdots,|\mathcal{Y}|\}$, where $|\mathcal{Y}|$ is total cardinality of label space.
    Then, our goal is to build a classifier $f_{cls}:\mathcal{H}\mapsto\mathcal{Y}$ which maps encoded features to proper classes with an unlabeled dataset $\mathcal{D} = \{\bm{x}_{i}\}_{i=1}^{k}$ and weak supervision signal (i.e., natural language name for each class).
    
\subsection{Prompt-tuning}
    \label{subsec:pl}
    Prompt-tuning projects an input sequence $x$ into a label space $\mathcal{Y}$ via LM's capability to fill the \texttt{\txtmagenta{[MASK]}} token.
    For instance, suppose we have to classify the binary sentiment (label 0: \texttt{positive}, label 1: \texttt{negative}) of the sentence.
    Given input sentence $\bm{x}=$ A great movie., we first transform the $\bm{x}$ into a cloze question through \textit{template}:
    \begin{equation*}
        \bm{x}^{\text{t}} = \pattern{\texttt{A \txtmagenta{[Mask]} review.}}\text{A great movie.}
    \end{equation*}
    Then we feed \textit{templified} input $\bm{x}^{\text{t}}$ to \texttt{LM} and extract the likelihood of the \texttt{\txtmagenta{[MASK]}} token.
    To convert the probability of extracted token into a label probability, we employ a \textit{verbalizer} $\mathcal{V}$, a few selected tokens from the whole vocabulary, and map them into corresponding label space $\mathcal{V}\mapsto\mathcal{Y}$.
    Specifically from the previous example, we can design a verbalizer utilizing a weak-supervision signal (i.e., name of each class) as follows:
    \begin{equation}
        \mathcal{V}_{0} = \{\texttt{positive}\}, \mathcal{V}_{1} = \{\texttt{negative}\}.
        \label{eq:verbalizer-baseline}
    \end{equation}
    Note that some recent studies have expanded the verbalizer words in Eq. \ref{eq:verbalizer-baseline} to multiple words by utilizing extra knowledge, such as ConceptNet \cite{speer2017conceptnet} or WordNet \cite{miller1995wordnet}, to make more accurate predictions.
    Then, for input $\bm{x}^{\text{t}}$, we measure the probability distribution of the label $P(y|\bm{x}^{\text{t}})$  utilizing verbalizer $\cup_{y\in\mathcal{Y}}\mathcal{V}_{y} = V$:
    \begin{equation}
        P(y|\bm{x}^{\text{t}}) =\frac{\sum_{w\in V_{y}}\texttt{LM}(\mask=w|\bm{x}^{\text{t}})}{\sum_{\mathcal{V}_{i}\in \mathcal{V}}\sum_{v\in V_{i}}\texttt{LM}(\mask=v|\bm{x}^{\text{t}})}
        \label{eq:verbalizer-prob}
    \end{equation}
    Finally, we compare the probability of $p(y=0)$ and $p(y=1)$ in the \texttt{\txtmagenta{[Mask]}} token and annotates input $\bm{x}$ with label 0.

\subsection{Linear Discriminative Analysis}
    \label{subsec:lda}
    LDA belongs to the generative classifier family which estimates the class probability of the input indirectly.
    Unlike discriminative classifier, which directly models the class posterior $p(y=c|\bm{x})$, LDA predicts $\bm{x}$ via bayes rule:
    \[
        p(y=c|\bm{x}) = \frac{p(\bm{x}|y=c) p(y=c)}{\Sigma_{c^{'}\in\mathcal{Y}}p(\bm{x}|y=c^{'}) p(y=c^{'})}.
    \]
    LDA assumes the class conditional densities follow multivariate Gaussian distribution with tied covariance over classes:
    \[
        p(\bm{x}|y=c) = \mathcal{N}(\bm{x}|\bm{\mu}_{c},\Sigma)
    \]
    Then the corresponding label probability (posterior) has the following form:
    \begin{equation*}
        \begin{split}
            p(y=c|\bm{x}) &\propto \pi_{c} \text{ }\mathcal{N}(\bm{x}|\bm{\mu}_{c},\Sigma)\\
            &\propto  \mathcal{N}(\bm{x}|\bm{\mu}_{c},\Sigma).
        \end{split}
    \end{equation*}
    The prior distribution $\pi_{c} = p(y=c)$ can be ignored as it is independent to $c$.

    \noindent \textbf{Train:} 
    To fit the LDA model, we have to estimate the Normal distribution of the target space $\mathcal{N}(\bm{\mu},\Sigma)$ from the available dataset $\mathcal{D}$. 
    Particularly, LDA estimates the distribution by employing MLE, which consists of empirical class mean $\bm{\hat{\mu}}_{c}$ and tied covariance $\hat{\Sigma}$.
        \[
            \bm{\hat{\mu}}_c=\frac{1}{|\mathcal{D}_c|}\sum_{n\in\mathcal{D}_c}\bm{x}_n,
        \]
        \[
            \hat{\Sigma}=\frac{1}{|\mathcal{D}|}\sum_{c=1}^{c\in\mathcal{Y}}\sum_{n\in\mathcal{D}_c} (\bm{x}_n-\bm{\hat{\mu}}_c)(\bm{x}_n-\bm{\hat{\mu}}_c)^{\texttt{T}}.
        \]

\begin{figure*}
    \begin{center}
        \includegraphics[width=0.99\textwidth]{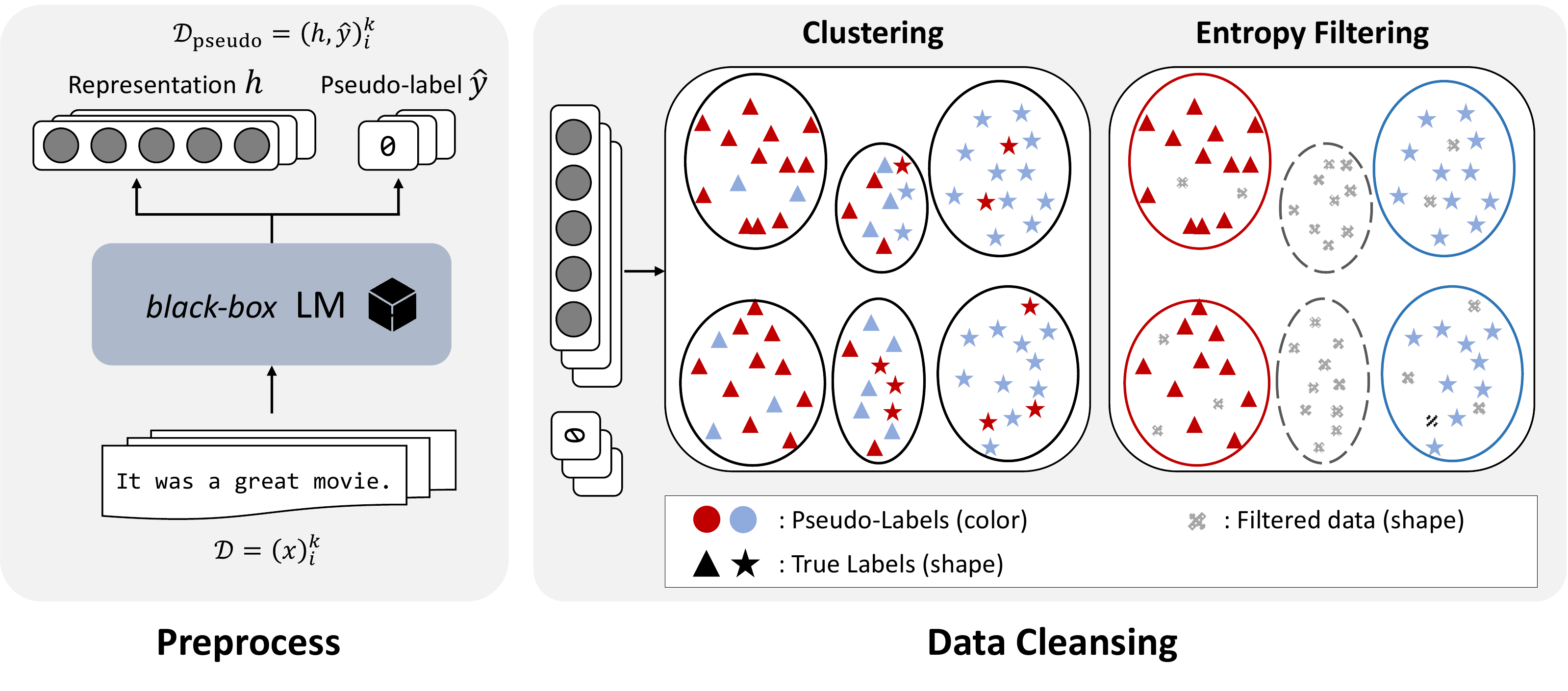}
          \caption{Illustration of \cref{subsec:pre}, \ref{subsec:dc} stage in \celda. We (1) extract pseudo label and latent feature pair from unlabeled data, and (2) discard some uncertain data points. }
          \label{fig:celda}
    \end{center}
\end{figure*}

    \noindent \textbf{Inference:} After training the trainable parameters, we can compute the probability of the class label through Mahalanobis distance $d_{\text{mah}}$ which measures the distance between the data point $\bm{x}$ and the estimated distribution $\mathcal{N}(\bm{\hat{\mu}},\hat{\Sigma})$:

    \begin{equation*}
        \begin{split}
            \hat{y} &= M(\bm{x}) = \argmax_{c}\log p(y=c|\bm{x}) \\
            & = \argmax_{c} \left[\log\pi_c - \frac{1}{2}d_{\text{mah}}(\bm{x},\bm{\mu}_c;\Sigma) + \text{C}\right]\\
            & = \argmin_{c} d_{\text{mah}}(\bm{x},\bm{\mu}_c;\Sigma),
        \end{split}
    \end{equation*}
    where $M$ refers to a trained LDA model and $d_{\text{mah}}(\bm{x},\bm{\mu}_c;\Sigma) = (\bm{x}_n-\bm{\hat{\mu}}_c)^{\texttt{T}}\Sigma^{-1}(\bm{x}_n-\bm{\hat{\mu}}_c)$.
    \\

%% file: Sections/04.Methods.tex
\section{\celda}
\label{sec:celda}
    We introduce \celda, Clustering-enhanced Linear Discriminative Analysis which enhances the text classification ability of \textit{black-box} LMs without labels.
    \celda\ first (1) consists pseudo labeled dataset by passing an unlabeled dataset to \texttt{LM}($\cdot$) and extracts high-quality representation and pseudo-label pair as introduced in \cref{subsec:pl}.
    Then, we (2) refine the pseudo labeled dataset into a small subset dataset dubbed a \textbf{certain dataset} by leveraging clustering and the concept of entropy. (Figure \ref{fig:celda} illustrates the (1), (2) stage of \celda\ training.)
    Finally, (3) we recursively train a third-party LDA module stacked on LM with the \textbf{certain dataset}, which learns a precise decision boundary from the noisy dataset.
    
    \subsection{Pre-processing}
        \label{subsec:pre}
        \celda\  first passes the unlabeled dataset $\mathcal{D} = \{\bm{x}_{i}\}_{i=1}^{k}$ to LM to extract the information required to train LDA.
        Specifically, we pass \textit{templified} sentence $\bm{x}^{\text{t}}$ to \texttt{LM}($\cdot$) and extract $\bm{h}_{\text{last}}$, $\bm{h}_{\text{verb}}$, and $\hat{y}$.
        Each notation refers to a mean-pooled last layer hidden representation $\bm{h}_{\text{last}}$, the verbalizer probability distribution of the \texttt{\txtmagenta{[Mask]}} token $\bm{h}_{\text{verb}}$, and its corresponding pseudo-label $\hat{y}$.
        \celda\  employs two different representations the $\bm{h}_{\text{last}}$ and $\bm{h}_{\text{verb}}$ since they retain complementary information which might be beneficial in learning a precise decision boundary. 
        Namely, the former encapsulates rich information, including semantics and syntactic information, and the latter possesses specific probability distribution of \texttt{\txtmagenta{[Mask]}} token.
        To maximize the capability of $\bm{h}_{\text{verb}}$, we expand the verbalizer in Eq. \ref{eq:verbalizer-baseline} into predefined multiple words following \citet{hu2021knowledgeable}.
        Finally, we concatenate both representations to derive a single final representation $\bm{h}$:
        \begin{equation}
            \bm{h} = [ \lVert \bm{h}_{\text{last}}\rVert_{2} \oplus \bm{h}_{\text{verbal}}].
            \label{eq:findal-rep}
        \end{equation}
        We apply L-2 normalization to $\bm{h}_{\text{last}}$ before concatenation to synchronize the range with the representation provided by the verbalizer (ranging from 0 to 1).
        Finally, we construct the pseudo labeled dataset $\mathcal{D}^{\text{pseudo}} = \{\bm{h}_{i}, \hat{y}_{i}\}_{i=1}^{k},$ for the next training phase.
        After this pre-processing step, we no longer use the LM, which tremendously reduces computational cost in training.

    \subsection{Data Cleansing with Cluster Entropy}
        \label{subsec:dc}
        The objective of this training phase is to filter out the uncertain data samples in $\mathcal{D}^{\text{pseudo}}$ to create a more reliable dataset, $\mathcal{D}^{\text{clean}}$. 
        To achieve this goal, we utilize clustering on the representations $\bm{h}$ based on the previous findings that the LM's features are grouped with semantically similar sentences. 
        Specifically, we employ KMeans to cluster the pseudo-labeled dataset $\mathcal{D}^{\text{pseudo}}$ into $K$ clusters. And let $\bm{h}_{ij}$ be the representation belonging to the $i^{\text{th}}$ cluster $K_{i}$ and $j$ be the index of the sample in the cluster. 
        From each cluster, we estimate the pseudo-label probability distribution within each cluster $P_{k_{i}}(\mathcal{Y})$:
        \[
            \text{\qquad} P_{K_{i}}(\mathcal{Y}=y) = \frac{|\mathcal{S}|}{|K_{i}|},
        \]
        where $\mathcal{S} = \{s \in \hat{y}_{i} | \hat{y}_{i} = y \}$ and $i$, $y$ denotes cluster index and label index, respectively.
        
        Then we measure the entropy weight (EW) of each cluster to estimate their uncertainty:
        \begin{equation}
            \text{EW}(K_{i}) = \frac{1 - \text{NormEnt}(K_{i})}{\sum_{j\in K}(1-\text{NormEnt}(K_{j}))},
            \label{eq:EW}
        \end{equation}
        \[
            \text{NormEnt}(K_{i}) = - \frac{\sum_{l \in \mathcal{Y}} p_{K_{i}}(l)\log p_{K_{i}}(l)}{\log|C|}.
        \]
        EW increases when the samples within a cluster tend to have the same label, but decreases otherwise.
        By setting a threshold, $\tau$, we are able to select a portion of certain clusters $K^{\text{clean}}$ while removing several uncertain clusters.
        
        \begin{equation}
            K^{\text{clean}} = \{K_{i} \in K |\ \text{EW}(K_{i}) \ge \tau\}
            \label{eq:cleansing1}
        \end{equation}
        Furthermore, we eliminate samples in the cluster that do not conform to the majority pseudo labels within the cluster and create the final certain dataset $\mathcal{D}^{\text{clean}}$:
        \begin{equation}
          \mathcal{D}^{\text{clean}} = \{K_{i} \in K^{\text{clean}} | \hat{y}_{ij} = \argmax_{y\in \mathcal{Y}}P_{K_i}(\mathcal{Y}) \}
            \label{eq:cleansing2}
        \end{equation}
        And we utilize the $\mathcal{D}^{\text{clean}}$ to train LDA.
        Figure \ref{fig:celda} illustrates \cref{subsec:pre}, \ref{subsec:dc} stage of \celda.

    \subsection{LDA Training}
    Utilizing the filtered dataset $\mathcal{D}^{\text{clean}}$ from prior step, we finally fit the parameters of LDA through MLE as introduced in \cref{subsec:lda}.
    To further improve LDA, we recursively train LDA by updating the pseudo labels with the trained model and repeating the previous data cleansing steps based on the assumption that the trained LDA produces more precise pseudo labels.
    To prevent the model from oscillating or diverging in iterative training, we employ moving average (MA):
    \[
       \bm{\hat{\mu}}_{t+1} = \frac{t-1}{t}\bm{\hat{\mu}}_{t-1} + \frac{1}{t}\bm{\hat{\mu}}_{t},
    \]
    \[
       \hat{\Sigma}_{t+1} = \frac{t-1}{t}\hat{\Sigma}_{t-1} + \frac{1}{t}\hat{\Sigma}_{t},
    \]
    where $t$ indicates timestamp (current epoch). 
    If the updated label does not deviate beyond a specific ratio $\delta$ from the previous label, we terminate the training process judging the model has converged.
    The overall procedure of \celda\ is summarized in Algorithm \ref{alg:train}.

\begin{algorithm}[t]
\caption{\celda\  Training.}
\label{alg:train}
\KwIn{
unlabeled dataset $\mathcal{D}$, a language model \texttt{LM}, pre-defined verbalizer $\mathcal{V}$, LDA model $M$
}
\textbf{Step 1}: Pre-processing

$\mathcal{D}^{\text{p}}=\{\}$

\For{$\bm{x}\in\mathcal{D}$}{
    {
    \text{\ \ 1a. get embedding} $\bm{h}$ $\gets$ \text{Eq.}\ref{eq:findal-rep};
    \text{2a. get pseudo-label} $\hat{y}$ $\gets$  \text{Eq.}\ref{eq:verbalizer-prob};
    $\mathcal{D}^{\text{p}}=\mathcal{D}^{\text{p}}\cup(\bm{h},\hat{y})$;\\
    }
}
Return $\mathcal{D}^p$\;

\textbf{Step 2}: Data Cleansing \& LDA training

\For{\text{until convergence.}}
{
    \text{1b. run KMeans on $\mathcal{D}^p$}  with $k=K$ until converge;\\
    \text{2b. data cleansing} $\mathcal{D}^{c}$ $\gets$ \text{Eq.}\ref{eq:cleansing1}, \ref{eq:cleansing2};\\
    \text{3b. train $M$ with} $\mathcal{D}^{c}$;\\
    \For{$(\bm{h},\hat{y}) \in \mathcal{D}^{p}$}{
        \text{4b. $\hat{y} = M(\bm{h}) \gets$update label} ;
    }

}
 
\KwOut{A trained LDA classifier $M$.}
\label{alg:celda}
\end{algorithm}

    As previously discussed in \cref{subsec:lda}, LDA is a variation of Gaussian Discriminative Analysis (GDA) that utilizes shared covariance among classes, resulting in a reduction of parameters from $O(\mathcal{|Y|}d^2)$ to $O(\mathcal{|Y|}d)$. 
    Despite the potential negative impact on performance compared to GDA when data is abundant and clean, LDA possesses several beneficial properties, such as the ability to fit the model with fewer samples and greater robustness to noisy labels \cite{murphy2022probabilistic}.
    Thus, the reduced parameters in LDA prevent overfitting and improve the model's ability to adapt to the test dataset, in contrast to other Gaussian-based approaches that are prone to overfitting and yields a poor performance in test cases, as will be demonstrated in our following experiments.
    The mentioned characteristics of LDA are highly valuable in our scenario, where the training dataset is noisy and a portion of the available data is discarded during the previous data cleansing stage. 

%% file: Sections/05.Experiments.tex
\section{Experiments}

\subsection{Backbone \& Datasets}
    We adopt T5 \cite{raffel2020exploring} as the main backbone of our experiments which is fairly large (up-to 11 billion parameters) and open-sourced. 
    Furthermore, we report additional experimental results with  SimCSE \cite{gao2021simcse} supervised RoBERTa-large \cite{liu2019roberta} in the Appendix.
    To investigate the performance of each method in many different scenarios, we carefully select 8 datasets \footnote{The detailed descriptions and references of each dataset are stipulated in the Appendix.}: AGNews, DBPedia, IMDb, SST2, Amazon, Yahoo, Banking77, and CLINC.
    The statistics of datasets used in the experiments are reported in Table \ref{tab:dset}. 
    To evaluate each method in stable conditions, we report the average accuracy of 5 different seeds (13, 27, 250, 583, 915) along with the corresponding standard deviation of 5 runs as a model performance.

\subsection{Experimental  Configurations}
    For all experiments, we utilize KMeans clustering with \textit{euclidean} distance, and set the number of clusters to $K = |\mathcal{Y}| \times 16$ except binary classification task.
    For binary classification tasks, we encountered an issue where the number of clusters became significantly smaller compared to the size of the total dataset. 
    As a solution, we opted to use a relatively large value for $K = |\mathcal{Y}| \times 64$.
    Furthermore, we set maximum tokenizer length and exit threshold $\delta$ adaptively depending on the dataset size.
    And we utilized the same template and verbalizer for every task from Openprompt \cite{ding2021openprompt}.
    Detailed configurations (i.e., templates, verbalizer) and our computation environments are stipulated in the Appendix.

\begin{table}[t]
\setlength{\tabcolsep}{9pt} 
\centering
\resizebox{\columnwidth}{!}{
\begin{tabular}{l|rrc}

\toprule
\textbf{Datasets} & \multicolumn{1}{c}{\textbf{\# Train}} & \multicolumn{1}{c}{\textbf{\# Test}} & \multicolumn{1}{c}{\textbf{\# Cls}} \\
\midrule
DBPedia & 560,000 & 70,000 & 14 \\
Yahoo & 1,400,000 & 60,000 & 10 \\
AGNews & 120,000 & 7,600 & 4 \\
SST2 & 6,920 & 1,821 & 2 \\
Amazon & 3,600,000 & 400,000 & 2 \\
IMDb & 25,000 & 25,000 & 2 \\
Banking77 & 10,003 & 3,080 & 77 \\
CLINC & 15,250 & 550 & 150 \\
\bottomrule
\end{tabular}
}
\caption{Dataset statistics}
\label{tab:dset}
\end{table}

\begin{table*}[t]
    \centering
    \resizebox{0.99\textwidth}{!}{
        \begin{tabular}{c|c|cccccc}
    \toprule
         \multicolumn{8}{c}{\textbf{T5-Large (770M)}} \\
    \midrule
        \textbf{Setting} & \textbf{Method} & \textbf{DBPedia (14)} & \textbf{Yahoo (10)} & \textbf{AGNews (4)} & \textbf{SST2 (2)} & \textbf{Amazon (2)} & \textbf{IMDb (2)} \\
    \midrule
        Full & LDA & 98.72 & 72.40 & 91.37 & 92.09 & 96.05 & 94.28 \\
    \cmidrule{1-2}
        \multirow{2}{*}{ZSL} & PET & 62.14 \footnotesize ± 0.0 & 34.00 \footnotesize ± 0.0 & 54.37 \footnotesize ± 0.0 & 69.96 \footnotesize ± 0.0 & 78.47 \footnotesize ± 0.1 & 77.96 \footnotesize ± 0.0 \\
         & KPT & 81.69 \footnotesize ± 0.6 & 62.32 \footnotesize ± 0.4 & 84.79 \footnotesize ± 0.8 & 82.70 \footnotesize ± 0.4 & 87.22 \footnotesize ± 0.7 & 83.71 \footnotesize ± 0.6 \\
    \cmidrule{1-2}
        \multirow{2}{*}{WSL} & SimPTC & 68.49 \footnotesize ± 0.1 & 47.9 \footnotesize ± 0.1 & 87.01 \footnotesize ± 0.0 & 68.48 \footnotesize ± 0.0 & \textbf{94.75} \footnotesize ± 0.1 & \textbf{92.74} \footnotesize ± 0.0 \\
         & \textsc{CeLDA} & \textbf{84.47} \footnotesize ± 0.3 & \textbf{68.88} \footnotesize ± 0.1 & \textbf{90.03} \footnotesize ± 0.2 & \textbf{89.40} \footnotesize ± 0.5 & 94.70 \footnotesize ± 0.1 & 91.70 \footnotesize ± 0.1 \\
    \midrule
        \multicolumn{8}{c}{\textbf{T5 (3B)}} \\
    \midrule
        Full & LDA & 98.75 & 73.73 & 92.11 & 94.23 & 96.74 & 95.35 \\
    \cmidrule{1-2}
        \multirow{2}{*}{ZSL} & PET & 62.11 \footnotesize ± 0.0 & 32.07 \footnotesize ± 0.0 & 46.36 \footnotesize ± 0.0 & 70.29 \footnotesize ± 0.0 & 68.84 \footnotesize ± 0.1 & 77.38 \footnotesize ± 0.0 \\
         & KPT & 82.78 \footnotesize ± 0.2 & 62.17 \footnotesize ± 0.2 & 86.03 \footnotesize ± 0.3 & 86.05 \footnotesize ± 0.1 & 84.15 \footnotesize ± 0.5 & 81.93 \footnotesize ± 1.3 \\
    \cmidrule{1-2}
        \multirow{2}{*}{WSL}  & SimPTC & 68.60 \footnotesize ± 0.2 & 50.20 \footnotesize ± 0.3 & 87.54 \footnotesize ± 0.1 & 89.79\footnotesize ± 0.1 & 95.99 \footnotesize ± 0.1 & 94.46 \footnotesize ± 0.0 \\
         & \celda & \textbf{85.11} \footnotesize ± 1.5 & \textbf{70.35} \footnotesize ± 0.1 & \textbf{90.18} \footnotesize ± 0.3 & \textbf{92.42} \footnotesize ± 0.1 & \textbf{96.08} \footnotesize ± 0.1& \textbf{94.55} \footnotesize ± 0.0 \\
        \midrule
        \multicolumn{8}{c}{\textbf{T5 (11B)}} \\
    \midrule
        Full & LDA & 98.87 & 73.64 & 92.21 & 94.51 & 97.03 & 95.88 \\
    \cmidrule{1-2}
        \multirow{2}{*}{ZSL} & PET & 65.47 \footnotesize ± 0.0 & 39.66 \footnotesize ± 0.0 & 62.51 \footnotesize ± 0.0 & 71.00 \footnotesize ± 0.0 & 71.58 \footnotesize ± 0.1 & 71.78 \footnotesize ± 0.0 \\
         & KPT & 83.45 \footnotesize ± 1.0 & 63.34 \footnotesize ± 0.2 & 86.09 \footnotesize ± 0.6 & 86.46 \footnotesize ± 0.1 & 88.03\footnotesize ± 0.4 & 84.08 \footnotesize ± 1.2 \\
    \cmidrule{1-2}
         \multirow{2}{*}{WSL} & SimPTC & 69.83 \footnotesize ± 0.1 & 51.02 \footnotesize ± 0.5 & 88.61 \footnotesize ± 0.1 & 88.58 \footnotesize ± 0.0 & 96.42 \footnotesize ± 0.0 & 95.02 \footnotesize ± 0.0 \\
         & \celda & \textbf{86.88} \footnotesize ± 0.3 & \textbf{71.38} \footnotesize ± 0.6 & \textbf{90.29} \footnotesize ± 0.3 & \textbf{94.23} \footnotesize ± 0.4 & \textbf{96.78} \footnotesize ± 0.0 & \textbf{95.61} \footnotesize ± 0.0 \\
    \bottomrule
        \end{tabular}
    }
    \caption{Experimental results on 6 benchmarks. Best method (except full LDA) for each dataset is indicated in \textbf{bold}. Full, ZSL, and WSL indicates fully supervised, zero-shot learning, and weakly-supervised learning, respectively. Our methods surpasses other compared methods and close the gap with fully supervised method.}
    \label{tab:main}
\end{table*}

\subsection{Competing Methods}
We compare our methods with state-of-the-art zero-shot text classification methods and several baseline fully supervised methods:

    \begin{itemize}
        \item \textbf{LDA (full)}:  We report the accuracy of the baseline LDA model trained on the top of LM representations with a fully-annotated dataset, which serve as our upper bound. 
        \item \textbf{PET}: Pattern-Exploiting Training \cite{schick2020exploiting} is a baseline zero-shot prompting method that transforms an input into a cloze-task and utilizes a single-word verbalizer to label data sample. 
        \item \textbf{KPT}: Knowledge Prompt-Tuning (KPT) \cite{hu-etal-2022-knowledgeable} is a multi-word verbalizer expansion of PET that extracts multi-words similar to the original label name from external knowledge bases.
        \item \textbf{SimPTC}: \cite{Fei2022BeyondPM} iteratively trains a Gaussian Mixture Model (GMM) on top of the LM's representations from the initial pseudo labels with E\&M algorithm, which measures the similarity between the anchor embedding and data sample.
    \end{itemize}

\subsection{Main Results}
Table \ref{tab:main} reports the performance of \celda\ and other competing methods on 6 benchmarks with 3 LMs of varying size (770M to 11B).
From the results, we share following observations:

\noindent \textbf{(1) Significant performance of \celda:} 
Our method consistently outperforms other competing methods by a large margin in varying language models.
Additionally, another black-box WSL method (SimPTC) often fails to reach a stable convergence point and performance drops as the training continues\footnote{Experimental setup in SimPTC paper utilizes both train and test dataset in training stage, unlike ours.}.
This is because they estimate the class-specialized covariance matrix resulting in exponential growth of the model parameter.
Thus SimPTC's decision boundary does not align well with the test distribution meaning that the model is prone to overfitting.
We delve into this phenomenon in detail in the Appendix.
However, our method constantly exhibits better performance and can capture precise decision boundaries without overfitting.
An additional noteworthy observation is that \celda\ approaches the fully supervised LDA with a small spread on most benchmarks.

\noindent \textbf{(2) Scalability of \celda to large models}:
Additionally, our method demonstrates scalability to large models, meaning the performance of \celda\  improves as the scale of the LM grows.
Other methods are not scalable since the performance of ZSL method easily saturates and displays insignificant improvement, and other well-performing WSL methods often require explicit model training.
However, the performance of \celda\ method improves with the larger language model, which indicates that our method exploits the high quality representations from the larger model quite properly.

%% file: Sections/06.Analysis.tex
\begin{figure*}[t]
    \centering
    \begin{subfigure}{.49\linewidth}
        \centering
        \includegraphics[width=.99\linewidth]{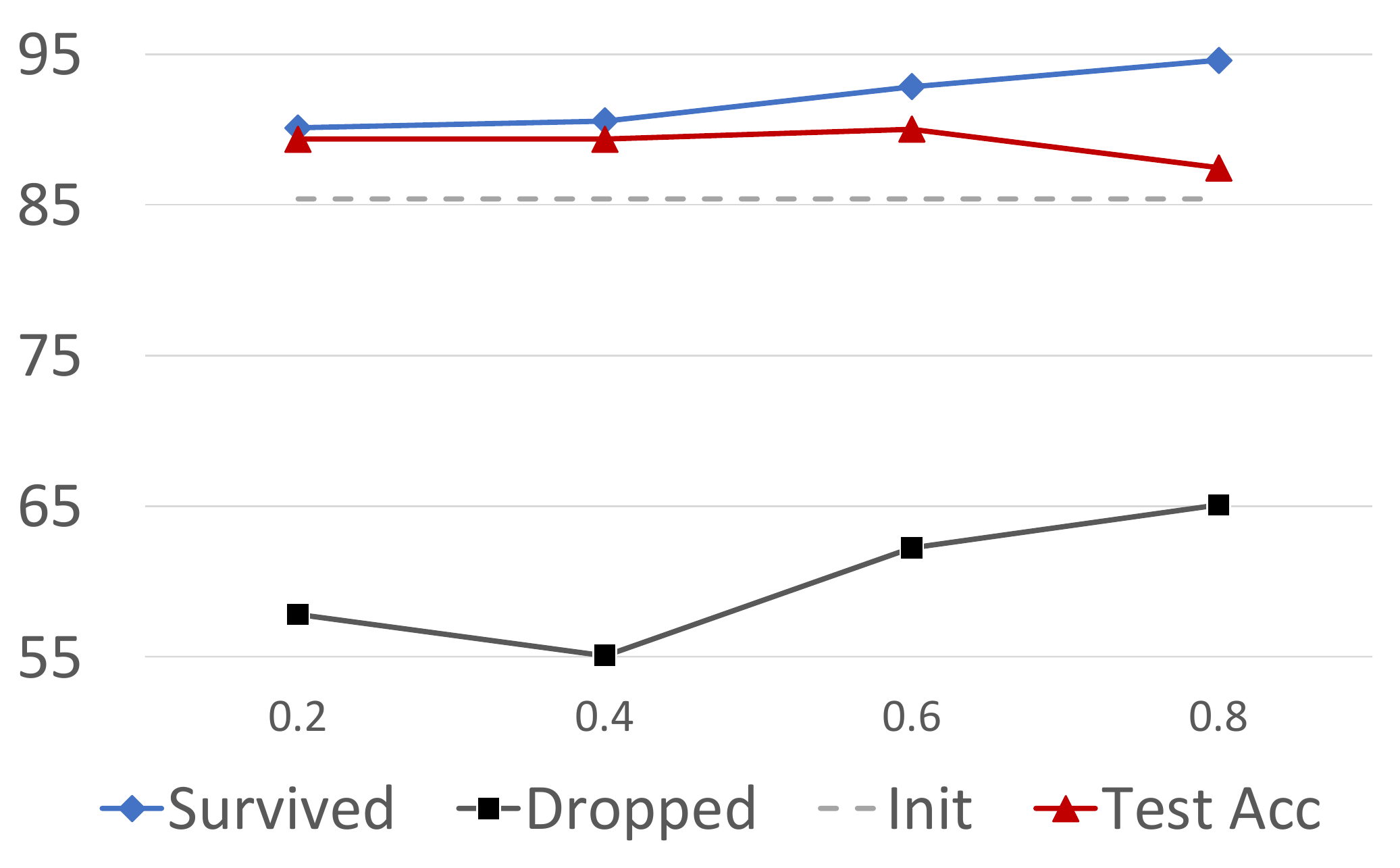}
        \caption{Accuracy of filtered \& abandoned datasets.}
        \label{fig:ab-filteracc}
    \end{subfigure}
    \begin{subfigure}{.49\linewidth}
        \centering
        \includegraphics[width=.99\linewidth]{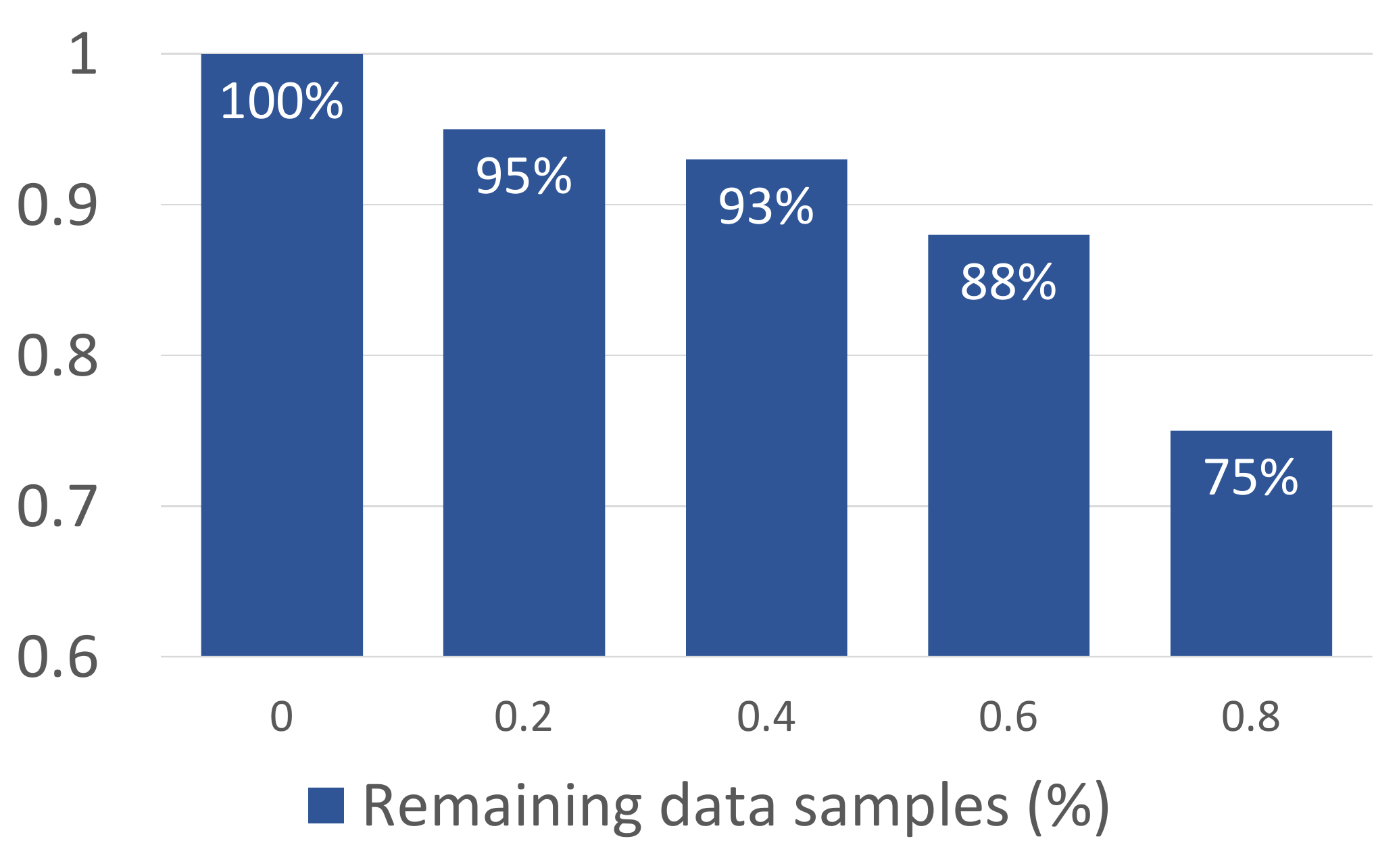}
        \caption{Remaining portion of dataset after filtering.}
        \label{fig:ab-numdset}
    \end{subfigure}
    \par \medskip
    \caption{Filtered dataset has more accurate data samples while the accuracy of dropped samples exhibits poor performance. Setting a high strong threshold increases the pseudo label accuracy but the overall performance drops.}
    \label{fig:ab1}
\end{figure*}

\section{Analysis}
    We conduct an in-depth investigation of \celda\ to elucidate its underlying mechanism in conjunction with our intuitions. 
    For all analysis experiments, we utilize T5-large as a backbone model.

\begin{table}[t]
\centering
\begin{tabular}{l|c}
\toprule
\multicolumn{1}{c|}{Model} & Accuracy \\
\midrule
Vanilla Prompting (PET) &  80.96\\
\; + LDA & 84.49 \\
\; \; + Recursive train & 85.45 \\
\; \; \; + multi-verbalizer & 88.10 \\
\; \; \; \; + feature augmentation & 88.57 \\
\midrule
\; \; \; \; \; + data cleansing (\celda) & \textbf{90.03} \\
\bottomrule
\end{tabular}
\caption{Component-wise ablation study in \celda}
\label{tab:component-ablation}
\end{table}

\subsection{Ablations}
    \label{sec:ablation}
    We carry out ablation studies in several aspects to further explore the effectiveness of the main components in our approach:
    
    \noindent \textbf{Component ablation:} 
    To validate our model design, we conducted a component-wise ablation study on the AGNews dataset, where we sequentially added each component in \celda. 
    From Table \ref{tab:component-ablation}, we confirmed that our components improve the overall performance progressively.
    Specifically, as mentioned \cref{subsec:lda}, LDA learns a more precise decision from a noisy labeled dataset, and recursive training also gives marginal improvement.
    Moreover, utilizing a multi-word verbalizer as in \citet{hu-etal-2022-knowledgeable} and augmenting their logit representation increases the overall performance.
    Finally, applying our clustering-based data cleansing approach improves the performance further and yields the best performance.
    
    \noindent \textbf{Effectiveness of data cleansing:}
    We scrutinize our clustering-based filtering module in various aspects to verify its efficacy.
    Firstly, we compare the quality of both datasets: One that has survived after the data filtering stage (survived dataset for abbreviation) and the other that has been dropped (dropped dataset for short).
    Figure \ref{fig:ab-filteracc} illustrates the accuracy when the cluster filtering threshold $\tau$ changes.
    From the figure, we verified that our survived dataset has more accurate data samples, while the accuracy of dropped samples lags far behind survived dataset.
    Moreover, setting a stronger threshold $\tau$ increases the pseudo-label accuracy of the survived dataset and reaches a near-clean dataset, but the overall performance drops.
    It implies that giving strong conditions discards even meaningful samples, which is beneficial in training.
    As a support, we can verify that the accuracy of the dropped dataset also increases with a higher $\tau$, and the total number of the survived sample decreases, as shown in Figure \ref{fig:ab-numdset}.

\begin{figure}[t]
    \begin{center}
        \includegraphics[width=1\columnwidth]{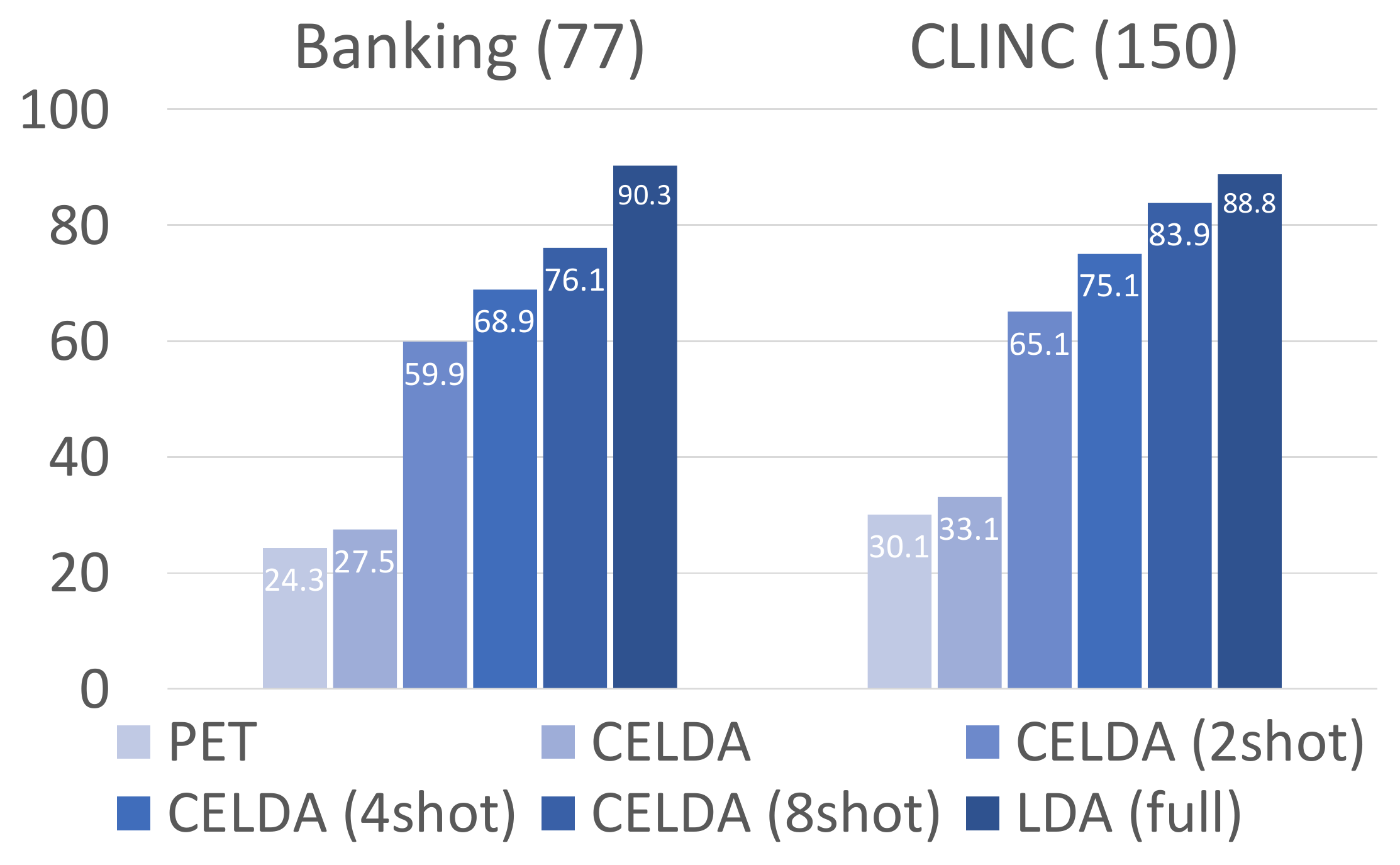}
          \caption{Experiments on fine-grained datasets. With a few true labels, our methods improves drastically. }
          \label{fig:ablation-al}
    \end{center}
\end{figure}

\subsection{Impact of Initial Pseudo-labels}
    Similar to other WSL methods, our methodology is heavily influenced by initial pseudo-labels.
    LMs generate highly reliable pseudo labels for coarsely labeled datasets, as shown in Figure \ref{tab:main}. 
    However, labeling a fine-grained dataset with LM often leads to poor performance, even with recent ZSL methods such as \cite{wang2021x, meng2020text}, only taking laborious manual engineering.
    Meanwhile, the hidden representations from LM have the potential to discriminate this fine-grained dataset without adaptation.
    For instance, in Banking dataset (77 classes) and CLINC dataset (150 classes), existing zero-shot labeling methods and WSL methods, including \celda, output unsatisfactory performances as seen in \ref{fig:ablation-al}.
    
    \celda\ can address this limitation by incorporate concepts from active learning (AL) which annotates a few selected samples with true labels through a human-in-the-loop pipeline.
    Specifically, we annotate $n$-shot samples per class (total $n \times |\mathcal{Y}|$ samples) that are closest to each centroid.
    Then, we annotate whole sentences in each cluster with the true label from the closest sample and re-train LDA as usual. 
    While this selection approach is quite simple, it selects highly-meaningful samples from the unlabeled dataset, significantly improving the performance of \celda\ sharply with minimal human annotated samples.
    As depicted in Figure \ref{fig:ablation-al}, we revealed that labeling 8 shots per class on fine-grained datasets significantly enhances the performance (by nearly 50\% on average), where traditional methods tend to struggle.

\begin{figure}[t]
    \begin{center}
        \includegraphics[width=0.9\columnwidth]{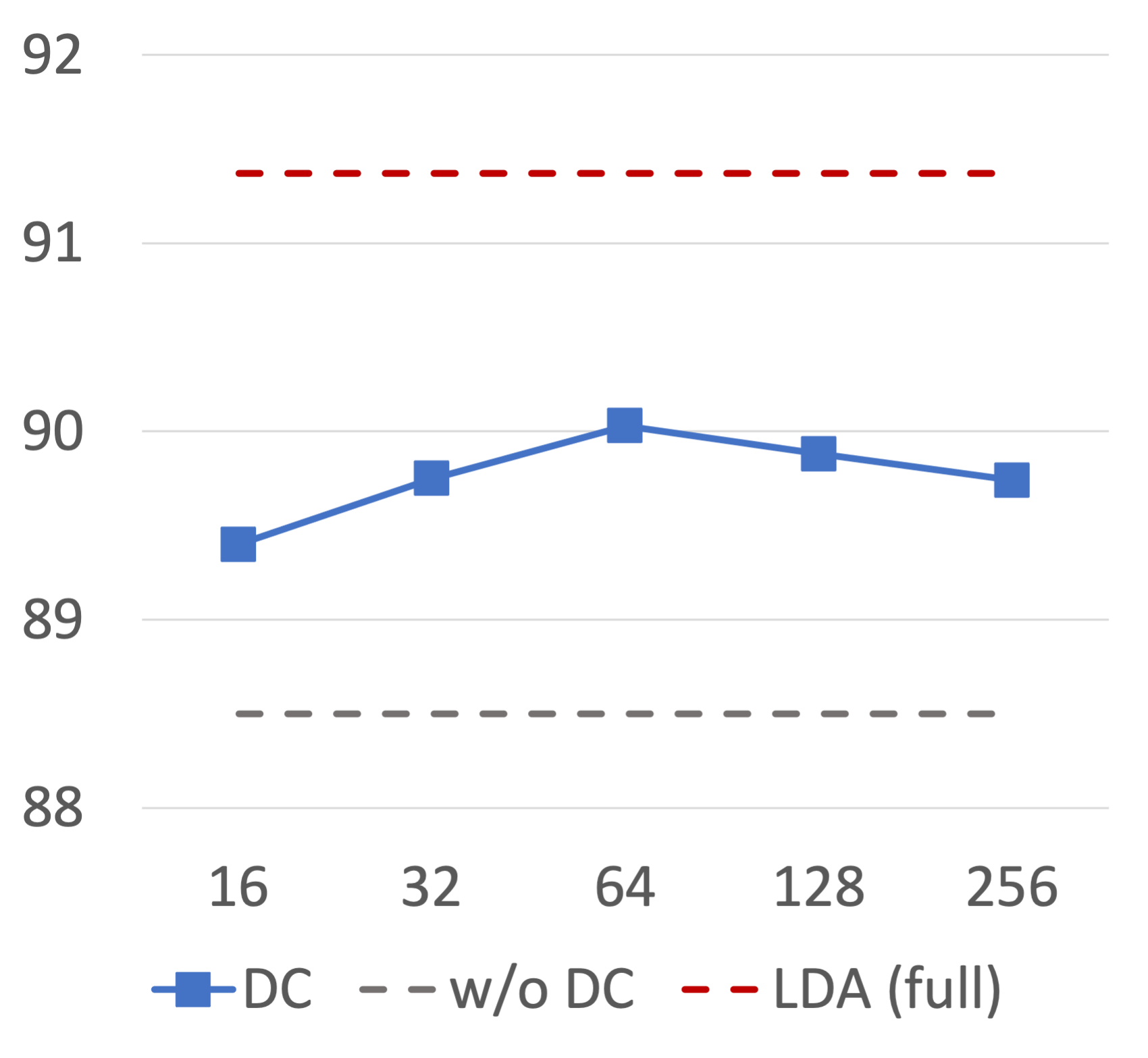}
          \caption{Correlation between the number of clusters and the performance.}
          \label{fig:ablation-clusternum}
    \end{center}
\end{figure}

    \subsection{Number of Clusters}
    We conducted an additional investigation of CELDA to analyze its the underlying mechanism. 
    We utilize T5-large as a backbone mode and tested on AGNews dataset.
    Figure \ref{fig:ablation-clusternum} illustrates the correlation between the number of clusters and the performance.
    We can confirm that the performance improves as the number of clusters increases, but slowly deteriorates when the number of clusters increases too much.
    As the total number of cluster increases, the samples belonging to each cluster decreases.
    Accordingly, entropy weight estimated from Eq. \ref{eq:EW} becomes unreliable hurting the effectiveness of the overall data cleansing process.
    Based on this result, we set number of clusters to $16\times|\mathcal{Y}|$.
    

%% file: Sections/07.Conclusion.tex
\section{Conclusion}
This work presents \celda, a practical framework for employing a black-box language model.
We have sought room for improvement in three orthogonal directions:
(1) Utilizing language models with high-quality representations (from the last layer and logit distribution).
(2) Filtering unreliable data samples from the noisy dataset.
(3) Recursively trains LDA, which is robust to noisy samples and avoids over-fitting by minimizing the overall model parameters.
By fusing these elements, we demonstrate the significant performance of \celda\ on sundry classification tasks and its scalability with the language model size.
In our follow-up study, we aim to employ sample-wise entropy from pseudo-labeling in the data cleansing phase instead of utilizing the entropy of the cluster, which is highly course-grained.
We expect that looking at fine-grained sample-wise entropy can yield a more precise data filtering effect, reducing meaningful samples from being dropped.

%% file: Sections/08.Limitations.tex
\section{Limitations}
    While our method demonstrates strong performance in our experimental setups, potential issues may arise when the characteristics of the available unlabeled dataset drastically change. 
    For one example, if the scale of the available dataset is too small, the effectiveness of our clustering-based data filtering may fall drastically, leading to poor performance.
    Or, if the dataset is highly unbalanced, our model cannot acquire information about several specific classes.
    One way to compensate for this shortcoming is to use an externally imported corpus or dataset, similar to other ZSL or WSL methods.
    Another drawback of \celda\ is that the final performance is highly dependent on the performance of the initial pseudo label, as shown in ablation.
    Nevertheless, as demonstrated in our ablation studies, we can remedy this issue by labeling a few samples, like active learning.

\section{Acknowledgement}
    This work was supported by Institute of Information \& communications Technology Planning \& Evaluation (IITP) grant funded by the Korea government(MSIT) [NO.2021-0-01343, Artificial Intelligence Graduate School Program (Seoul National University)] 

%% file: Sections/99.Appendix.tex
\appendix

\section*{Appendix}

\section{Dataset Description}
    In our experiments we use 8 different benchmark datasets which include topic classification, intent classification, question classification, and binary sentiment classification datasets.
    
    \noindent \textbf{DBpeida} \cite{brummer2016dbpedia} is an ontology classification dataset with DBpedia documents and 14 topics. It is a balanced dataset containing 40,000 training data and 5,000 testing data per class. 
      
    \noindent \textbf{Yahoo} \cite{zhang2015character} dataset is composed of a pair of questions and answers and a topic of it.
      
    \noindent \textbf{AGNews} \cite{zhang2015character} is a news topic classification dataset from AG’s news corpus with 4 different classes. 
      
    \noindent \textbf{IMDb} \cite{maas2011learning} is a binary movie review dataset for sentiment classification.
      
    \noindent \textbf{SST-2} \cite{socher2013recursive} is for detecting the sentiment of a single sentence of the movie review.
      
    \noindent \textbf{Amazon} \cite{mcauley2013hidden} is a review dataset Amazon from various domains (e.g., electronic stuff) for sentiment classification.
        
    \noindent \textbf{Banking77} \cite{casanueva2020efficient} a intent classification dataset which comprises fine-grained 77 intents in a single \textit{banking} domain regarding customer service queries.
        
    \noindent \textbf{CLINC} \cite{larson2019evaluation} is for classifying an intention of queries in dialog systems. The classes of CLINC dataset cover a total of 150 classes in 10 different domains and one out-of-scope class.

\section{Detailed Implementation Details}

    In pseudo label initialization of training data, a zero-shot prediction ability of T5 is utilized with a template and verbalizers. Expanded label words are used as verbalizers in experiments of main datasets for rich logit representations and pseudo labels. We also take a template with a mask to get mask logit values of pre-defined verbalizers from the pre-trained T5 model. 

    \noindent \textbf{Templates and Verbalizers}:
    We apply manual templates from OpenPrompt \cite{ding2021openprompt} in zero-shot pseudo-labeling which are listed in Table \ref{tab:templates}.
    To annotate samples with pseudo-labeling, expanded label words constructed by KPT \cite{hu2021knowledgeable} are employed for each task (see Table \ref{tab:verbalizer}). 
    For Banking77 and CLINC dataset, we use true label words without any expansion due to their abundant classes (77 and classes).

\label{sec:appendix}

\begin{table*}[t]
\setlength{\tabcolsep}{18pt} 
\centering
\resizebox{0.9\textwidth}{!}{
\begin{tabular}{c|l|l}
\toprule
\textbf{Task type} & \multicolumn{1}{c|}{\textbf{Dataset}} & \multicolumn{1}{c}{\textbf{Template}} \\
\midrule
Sentiment & IMDb, SST2, Amazon & It was \mask . {[}input sentence{]} \\
\midrule
\multirow{4}{*}{Topic} & DBPedia & {[}input sentence{]} is a \mask. \\
 & Yahoo & A \mask\ question : {[}input sentence{]} \\
 & AGNews & A \mask\ news : {[}input sentence{]} \\
 & Banking77, CLINC & {[} Category : \mask{]} {[}input sentence{]} \\
\bottomrule
\end{tabular}
}
\caption{Pre-defined templates for each task.}
\label{tab:templates}
\end{table*}

\begin{table*}[t]
\centering
\resizebox{0.99\textwidth}{!}{
\begin{tabular}{c|ll}
\toprule
\multicolumn{1}{c|}{\textbf{Dataset}} & \multicolumn{1}{c}{\textbf{Verbalizer (\txtmagenta{True label word}: Expanded words)}} &  \\
\midrule
\begin{tabular}[c]{@{}l@{}}IMDb, SST2, \\ Amazon\end{tabular} & \begin{tabular}[c]{@{}l@{}}\txtmagenta{negative}: bad,abysmal,adverse,alarming,angry,annoy, anxious, worthless,wound,yell,yucky, ...\\ \txtmagenta{positive}: good,absolutely,accepted,acclaimed,accomplish, wealthy,welcome,well,whole, ...\end{tabular} &  \\
\midrule
DBPedia & \begin{tabular}[c]{@{}l@{}}\txtmagenta{company}: corporation, company, corp, shareholder, enterprise, conglomerate, firm, ...\\ \txtmagenta{school}: school, education, university, academy, college, teacher, classroom, ...\\ ...\\ \txtmagenta{book}: novel, publication, book, fiction, publishing, author, prose, magazine, text, novella, ...\end{tabular} &  \\
\midrule
Yahoo & \begin{tabular}[c]{@{}l@{}}\txtmagenta{society}: society, culture, civilization, philosophy, association, anthropology, guild, subculture, ...\\ \txtmagenta{science}: science, mathematics, biology, mathematician, scientist, calculus, geometry, ...\\ ...\\ \txtmagenta{politics}: politics, government, governance, administration, law, democracy, aristotle, state, ...\end{tabular} &  \\
\midrule
AGNews & \begin{tabular}[c]{@{}l@{}}\txtmagenta{politics}: politics,government,diplomatic,law,aristotle,diplomatical,governance,republic, ...\\ \txtmagenta{sports}: sports,athletics,gymnastics,sportsman,competition,cycling,soccer,tennis,game, ...\\ ...\\ \txtmagenta{technology}: technology,engineering,science,biotechnology,internet,nanotechnology, ...\end{tabular} & \\
\bottomrule
\end{tabular}
}
\caption{Pre-defined verbalizers for each task.}
\label{tab:verbalizer}
\end{table*}

\begin{table*}[t]
\resizebox{1\textwidth}{!}{
\begin{tabular}{l|c|cccccc}
\toprule
 \textbf{Split} & \textbf{Method} & \textbf{DBPedia (14)} &\textbf{ Yahoo (10)} & \textbf{AGNews (4)} & \textbf{IMDb (2)} & \textbf{SST2 (2)} & \textbf{Amazon (2)} \\
 \midrule
 
\multirow{2}{*}{Train} & SimPTC & 67.22  \footnotesize ± 0.5 & 50.60  \footnotesize ± 0.1 & 85.53  \footnotesize ± 0.4 & \textbf{88.13}  \footnotesize ± 0.0 & 89.99  \footnotesize ± 0.3 & 94.46  \footnotesize ± 0.2 \\

 & CELDA & \textbf{83.43 } \footnotesize ± 0.1 & \textbf{61.34}  \footnotesize ± 0.9 & \textbf{88.82 } \footnotesize ± 0.5 & 87.11 \footnotesize ± 0.1& \textbf{90.18 } \footnotesize ± 0.2 & \textbf{94.50}  \footnotesize ± 0.1\\
 
 \midrule

\multirow{2}{*}{\; + Test} & SimPTC & 80.43  \footnotesize ± 0.0 & \textbf{63.13}  \footnotesize ± 0.0 & 85.82  \footnotesize ± 0.1 & \textbf{88.47}  \footnotesize ± 0.0 & 88.8  \footnotesize ± 0.2 & 94.38  \footnotesize ± 0.1 \\

 & CELDA & \textbf{83.47}  \footnotesize ± 0.2 & 62.62 \footnotesize ± 0.1 & \textbf{88.09}  \footnotesize ± 0.4 & 87.62\footnotesize ± 0.1 & \textbf{89.91}  \footnotesize ± 0.1 & \textbf{94.39}  \footnotesize ± 0.1 \\
 \bottomrule

\end{tabular}
}
\caption{Experimental results on 6 datasets with SimCSE-Roberta-Large. While our method constantly exhibits similar performance with or without a test split, SimPTC’s decision boundary does not align well without the test split, meaning that the model is prone to overfitting.}
\label{appen-tab:simcse-roberta}
\end{table*}

    \noindent \textbf{Environments and Utilized Libraries}:
    We utilize 8 RTX A6000 (48GB) GPUs for the experiments. When we extract initial pseudo labels and representations, KPT framework \cite{hu2021knowledgeable} in a zero-shot setting is used with OpenPrompt \cite{ding2021openprompt} library. Among various sizes of T5 models, we utilize T5-large, T5-3B, and T5-11B from Transformers \cite{wolf-etal-2020-transformers} library in Huggingface. The batch size is set adaptively depending on the average length of each dataset and the size of T5. KMeans code from \url{https://github.com/subhadarship/kmeans_pytorch} library is also used in CELDA.

\section{Comparison with SimPTC}
    As an implementation detail of SimPTC \cite{fei2022beyond}, both the train and test datasets are used in fitting Bayesian Gaussian Mixture Model (BGMM) by considering them as a set of unlabeled data. Then, SimPTC measures the accuracy of the test dataset, which is a portion of the unlabeled dataset used in training GMM. It is different from our setting of using only a train dataset to train LDA. Thus we additionally experiment with the setting of SimPTC. 
    
    Since SimPTC mainly uses SimCSE \cite{gao2021simcse} supervised RoBERTa large \cite{liu2019roberta} embeddings in their experiments, we also extract embeddings and construct pseudo labels of samples with SimCSE supervised RoBERTa large. SimCSE supervised RoBERTa significantly loses its ability of predicting masked words while fine-tuned with contrastive objective without MLM. Thus we could not perform mask prediction based zero-shot pseudo labeling with SimCSE supervised RoBERTa large. Instead, we initialize pseudo labels with Encode \& Match, a process of generating pseudo-label, proposed by SimPTC, which assign a pseudo-label to each input embedding with class anchor sentence embeddings. Consequently, initial pseudo labels of samples are identical for both CELDA and SimPTC.

    We follow SimPTC's design of the experiment in reproducing its performance. The experiment with CELDA is performed without the representation augmentation since verbalizer logit representation is unavailable. 
    
    According to the results in Table \ref{appen-tab:simcse-roberta}, CELDA outperforms SimPTC in most of the experiments. 
    Especially, CELDA displays a better accuracy in most cases when we utilize only a train split dataset which is a usual case in machine learning.
    Even in using both train and test datasets as an unlabeled dataset, CELDA outperforms SimPTC in most of the results. 
    While LDA in CELDA shares covariance among classes, GMM in SimPTC computes a covariance matrix for each class. 
    It causes overfitting on training samples and results in poor performance in test cases. For CELDA, the overall performance of two different settings are stable with a tied-covariance. Since SimPTC uses tied-covariance setting for IMDb and Amazon datasets, the performance of SimPTC and CELDA are close in both cases.